\pdfoutput=1
\documentclass{article}
\usepackage{microtype}
\usepackage{graphicx}
\usepackage{booktabs}
\usepackage{array}
\usepackage{adjustbox}
\usepackage{hyperref}
\usepackage[T1]{fontenc}
\usepackage[utf8]{inputenc}
\usepackage{textcomp}
\usepackage{xcolor}
\usepackage[most]{tcolorbox}

\usepackage[accepted]{icml2025}
\definecolor{nx}{HTML}{0B6E8A}
\newtcolorbox{findingbox}{colback=nx!7,colframe=nx,boxrule=0pt,leftrule=2pt,arc=0pt,left=5pt,right=5pt,top=3pt,bottom=3pt,fontupper=\small}
\icmltitlerunning{Nexus: Empowering Every Device with Safe On-Device AI}

\begin{document}
\hypersetup{pdftitle={Nexus: Empowering Every Device with Safe On-Device AI},pdfauthor={Pavan C Shekar, Aswanth Krishnan, Arpit Jain},pdfsubject={},pdfkeywords={on-device AI, model recommendation, quantization, edge deployment, safety}}
\twocolumn[
\icmltitle{Nexus: Empowering Every Device with Safe On-Device AI}
\begin{icmlauthorlist}
\icmlauthor{Pavan C Shekar}{qpiai}
\icmlauthor{Aswanth Krishnan}{qpiai}
\icmlauthor{Arpit Jain}{qpiai}
\end{icmlauthorlist}
\icmlaffiliation{qpiai}{QpiAI, Bengaluru, India}
\icmlcorrespondingauthor{Pavan C Shekar}{pavan.s@qpiai.tech}
\icmlcorrespondingauthor{Aswanth Krishnan}{ashwanth.krishnan@qpiai.tech}
\icmlcorrespondingauthor{Arpit Jain}{arpit.j@qpiai.tech}
\icmlkeywords{on-device AI, model recommendation, quantization, edge deployment, safety}
\vskip 0.3in
]
{\let\thefootnote\relax\footnotetext{\raggedright\textsuperscript{1}QpiAI, Bengaluru, India. Correspondence to: \mbox{Pavan C Shekar \textless{}\href{mailto:pavan.s@qpiai.tech}{pavan.s@qpiai.tech}\textgreater{}}, \mbox{Aswanth Krishnan \textless{}\href{mailto:ashwanth.krishnan@qpiai.tech}{ashwanth.krishnan@qpiai.tech}\textgreater{}}, \mbox{Arpit Jain \textless{}\href{mailto:arpit.j@qpiai.tech}{arpit.j@qpiai.tech}\textgreater{}}. Preprint, November 2026.}}

\begin{abstract}
Everyday devices such as phones, laptops and desktops can now run language models offline, but they are compute-constrained and typically have between 8 and 32 GB of memory, and a model that fits one device may not fit another. Choosing a model, a file format and a precision that together fit a given device is hard, and a poor choice could leave the device unrecoverable. Existing work improves single stages such as quantization or runtime speed, and tools that estimate memory fit never run the model. What is missing is an end-to-end path that checks safety before any ranking, delivers the chosen model, and measures it on the device. We build that path as Nexus and test it on one real phone, emulated small devices and a server: on the phone, a deployed model matched the server's pass or fail on 171 of 180 items, and every model marked safe on the devices loaded successfully. Code: \url{https://github.com/qpiai/nexus}.
\end{abstract}

\section{Introduction}\label{sec:intro}

We believe the benefits of AI should reach everyone. One way to do this is to make every device more intelligent and powerful: by empowering devices, we can empower the individuals who use them. Phones, laptops and desktop computers can already run language models locally, without a network connection and without sending data to a remote service.

\begin{figure*}[t]
\centering
\includegraphics[width=\linewidth]{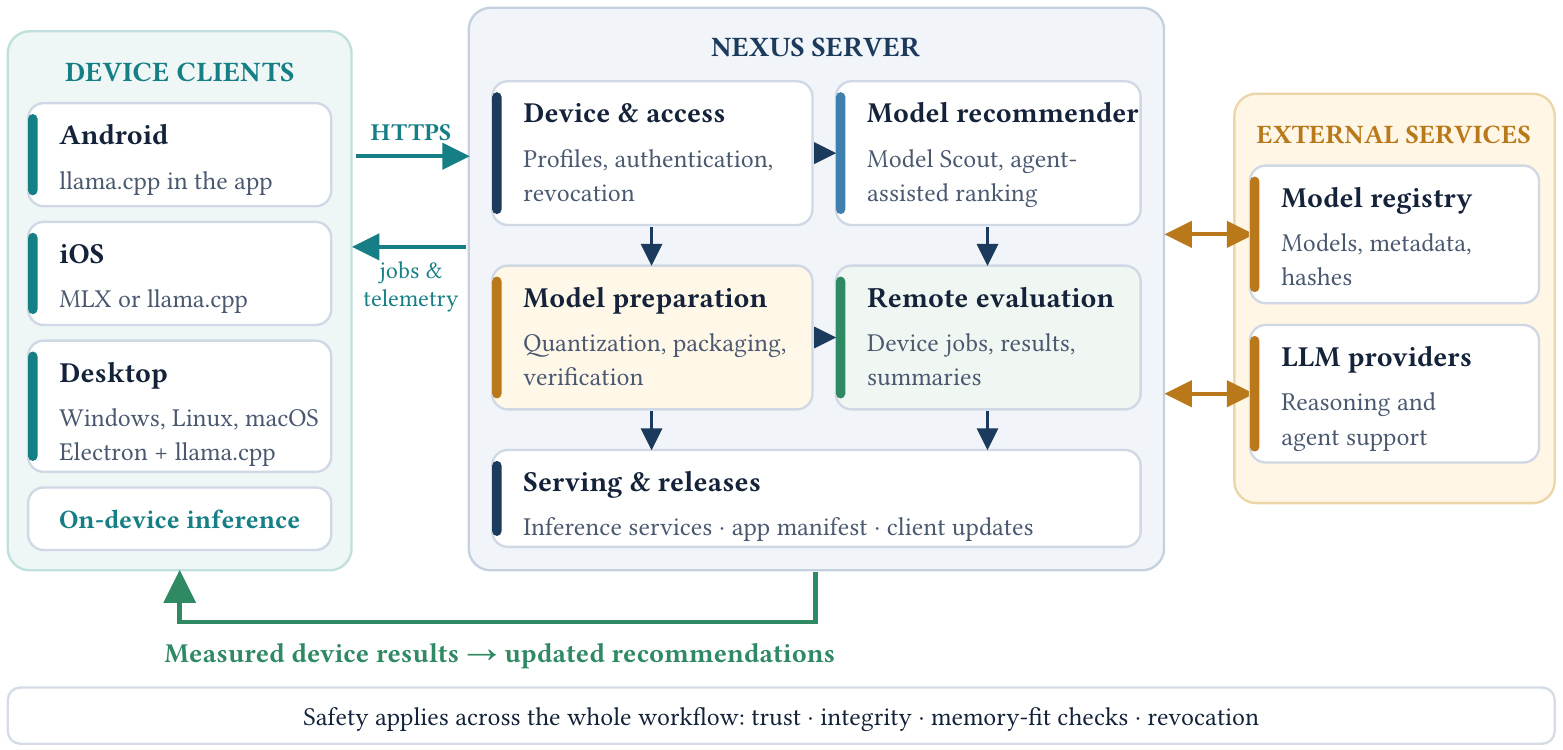}
\caption{Nexus architecture. Device clients (Android, iOS, desktop) run inference locally and exchange jobs and telemetry with the server over HTTPS. The server profiles devices, recommends and prepares models, and runs evaluation jobs on the devices; measured device results flow back into updated recommendations.}
\label{fig:arch}
\end{figure*}

\textbf{Devices are constrained.} A typical phone or laptop has 8, 16 or 32 GB of memory, and a GPU, when present, has 4 or 8 GB. Currently there are two trends, both good: small models keep becoming more capable, and devices keep getting more powerful hardware. For any one device, these trends meet at a \emph{Pareto frontier}: for each amount of memory, the best quality that any model and precision can reach (Figure~\ref{fig:frontier}). Even so, a model that does useful work still strains most everyday devices, and most users land well below this frontier, because reaching it requires choosing a model, a file format and a precision that together fit the device. This is a multi-variable optimization, informed by our experimental experience.

\textbf{The world is diverse.} Devices differ in brand, operating system, processor architecture and form factor, from phones and personal computers to edge and IoT hardware. No single model or runtime serves them all, so there is no one-size-fits-all solution, and one universal model cannot be run on every device.

\textbf{Safety is not optional.} A model that exceeds what a device can sustain does not merely fail to load; it can crash the device, lose data and force a replacement. In one of our early experiments, running Llama 2 7B~\cite{llama2} on a Samsung Galaxy M31s with 8 GB of memory left the phone unrecoverable, with damage to its processor and motherboard.

\textbf{What exists, and what is missing.} Existing work optimizes individual stages: quantization methods shrink models~\cite{gptq,awq}, runtimes execute them efficiently on commodity hardware~\cite{llamacpp,mlx,llmnpu}, and measurement studies characterize how they behave on phones~\cite{survey}. What a user needs is the whole path, from choosing a model to running it, with a guarantee that the result will not harm the device and a way to learn whether it actually works. We built Nexus to provide that path, end to end.

\textbf{Nexus in brief.} Nexus (Figure~\ref{fig:arch}) is a server and a set of native clients. A client reports the device's memory, storage and runtime; the server checks every candidate model against four deterministic gates, lets language-model agents rank only the models that pass, prepares the chosen model (quantization, packaging, verification), and delivers it to the client, which runs it on the device. The client then runs test jobs on its own copy of the model and reports speed, temperature and answer quality, which the server uses in later recommendations.

\textbf{Hypothesis.} Our central hypothesis is that deterministic safety checks placed before any learned decision, combined with evaluation on the device itself, can give every device the right, best and safest model it can run. We test it in order from the most realistic setting to the most controlled: a real phone, emulated small devices, and a server.

This paper makes four contributions:
\begin{itemize}
\item \textbf{A safety-first recommendation design}, in which deterministic gates decide which models are eligible and agents only rank among them, so an agent cannot recommend a model the checks have rejected (Section~\ref{sec:recommend}).
\item \textbf{An end-to-end system}, from model selection through quantization and fine-tuning to native clients on five platforms, each with a device-specific runtime (Section~\ref{sec:optimize}, Section~\ref{sec:clients}). Each client also checks a model against the device's current free memory before loading it, as a second safety layer after the server's gates.
\item \textbf{Device-side remote evaluation}, in which the server queues test jobs that devices run on their own copy of a model, turning deployment into a source of evidence (Section~\ref{sec:monitor}).
\item \textbf{An evaluation} on a real phone, emulated devices and a server (Section~\ref{sec:eval}, Section~\ref{sec:limits}).
\end{itemize}

\begin{table*}[t]
\caption{Findings at a glance. Evidence: finding 1 in Section~\ref{sec:phone}, 2 in Section~\ref{sec:fit} and 3 in Section~\ref{sec:compress}.}
\label{tab:glance}
\centering\small
\begin{adjustbox}{max width=\linewidth}
\begin{tabular}{>{\centering\arraybackslash}p{0.042\linewidth}>{\raggedright\arraybackslash}p{0.333\linewidth}>{\raggedright\arraybackslash}p{0.555\linewidth}}
\toprule
\# & Finding & Key evidence \\
\midrule
1 & Phone and server agree; heat and longer context slow the phone & 171/180 same pass or fail; 31.2 $\rightarrow$ 21.4 tok/s as battery warms 33 $\rightarrow$ 43 \textdegree{}C; 82\% of short-prompt speed at about 1,200 tokens of context \\
2 & In our tests, models marked safe always loaded, and the client guard never loaded an unsafe one & 17 of 17 ``fits'' loaded; client agreed in 15/15 cases; sub-1B top pick 64/64 $\rightarrow$ 24/64 \\
3 & Spend memory on parameters, down to about 3 bits & 1.7B at 3 bits (860 MB) perplexity 12.9 vs 16.9 for 360M at 16 bits (724 MB); calibration cuts D from 0.118 to 0.036 \\
\bottomrule
\end{tabular}
\end{adjustbox}
\end{table*}

\section{Related Work}\label{sec:related}

\textbf{On-device language models.} Running language models on phones and personal computers has become practical through small models and efficient runtimes. Surveys cover model compression, hardware acceleration and edge--cloud collaboration~\cite{survey}. llama.cpp~\cite{llamacpp} runs quantized models on commodity CPUs and GPUs through its GGUF file format, MLX~\cite{mlx} targets Apple silicon, and llm.npu~\cite{llmnpu} offloads the prompt phase to mobile NPUs. Small model families such as SmolLM2~\cite{smollm2}, Qwen3~\cite{qwen3}, Gemma 2~\cite{gemma2} and Phi-3~\cite{phi3} are designed to fit such devices, and MobileLLM~\cite{mobilellm} tailors the architecture of sub-billion-parameter models to on-device use. These systems make execution possible; Nexus decides \emph{which} model to execute on a given device and verifies the outcome.

\textbf{Choosing and testing models for a device.} Tools for running models locally have started to estimate fit. llmfit~\cite{llmfit} detects a machine's memory, processor and graphics card and scores catalog models for memory fit, estimated speed and quality, but its scores are estimates and it never runs the model. Benchmarking frameworks such as MELT~\cite{melt} and MobileAIBench~\cite{mobileaibench} run models on real phones and measure speed, memory, energy and accuracy, but they are measurement tools rather than recommenders, and their results do not flow back into a choice for a particular device.

\textbf{Quantization and precision.} Post-training quantization reduces weights to fewer bits. GPTQ~\cite{gptq} uses approximate second-order information, and AWQ~\cite{awq} protects the weights that matter most for activations; LLM.int8()~\cite{llmint8} and SmoothQuant~\cite{smoothquant} make 8-bit weights and activations practical by handling activation outliers; BitNet~\cite{bitnet} and its 1.58-bit successor~\cite{bitnet158} train with very low-bit weights from the start. Scaling studies show that, for a fixed memory budget, 4-bit precision with more parameters tends to outperform higher precision with fewer~\cite{kbit}, and that the trade-off depends on both training and inference precision~\cite{precision}. A systematic evaluation on commodity hardware found that heavily quantized large models consistently outperform smaller high-precision ones, down to about 3.5 effective bits per weight~\cite{song2025}. Our measurements on real llama.cpp files agree, and we use this result to drive recommendation decisions.

\textbf{Efficient inference and adaptation.} Generating text on memory-bound hardware is limited by the rate at which weights are read, as the roofline model describes~\cite{roofline}; batching several requests shares each read~\cite{orca,pagedattn}, memory-efficient attention kernels reduce the memory traffic of attention~\cite{flashattn}, and speculative decoding reduces the number of sequential steps~\cite{specdec}. Low-rank adapters~\cite{lora} and their quantized variant~\cite{qlora} fine-tune a model by training a small number of added parameters.

Prior work optimizes one stage at a time and evaluates it in isolation. Nexus composes these stages into one loop, places deterministic safety checks before any learned component, and uses the deployed device as a measurement instrument; to the best of our knowledge, this combination has not been done before.

\section{Design Goals}\label{sec:goals}

We designed Nexus around four goals that follow from the problems above.
\begin{enumerate}
\item \textbf{Safety comes first.} No recommendation or delivery may exceed what a device can sustain. The checks that enforce this must be deterministic and must run before any learned component can act.

\item \textbf{Diversity is handled by design.} The system serves different operating systems, processors and form factors with a runtime suited to each, rather than one universal model.

\item \textbf{Best use of limited resources.} Within the safety limits, each device should receive the best model and precision it can run: the best option given the device's constraints and trade-offs.

\item \textbf{The system evolves.} A model that fits today may not fit tomorrow: a device can lose capability (it runs hot, its battery wears, other apps take its memory and storage), and newer models and better methods appear. Nexus keeps measuring and can change its recommendation.
\end{enumerate}

\section{System Design}\label{sec:system}

\subsection{Overview}\label{sec:overview}

Figure~\ref{fig:arch} shows the architecture. Device clients communicate with the Nexus server over HTTPS. The server profiles and authenticates devices, recommends models, prepares them through quantization, packaging and verification, and runs remote evaluations as jobs on the devices themselves. A serving and release layer distributes inference services, the app manifest and client updates. External services provide the model registry (model metadata and hashes) and the language-model providers that support the reasoning agents. Safety checks for trust, integrity, memory fit and revocation apply across the whole workflow.

Implementation notes. The server is a web application with a device registry, the recommender, a model scout that reads live listings from Hugging Face, a queue of remote jobs, a release manifest, and workers that quantize models. A pool of llama.cpp server processes answers chat requests. The clients wrap llama.cpp: the Android app has it built in, the desktop app runs it inside Electron, and the iOS app through MLX or llama.cpp.

\subsection{Safety-Aware Model Recommendation}\label{sec:recommend}

\begin{figure*}[t]
\centering
\includegraphics[width=\linewidth]{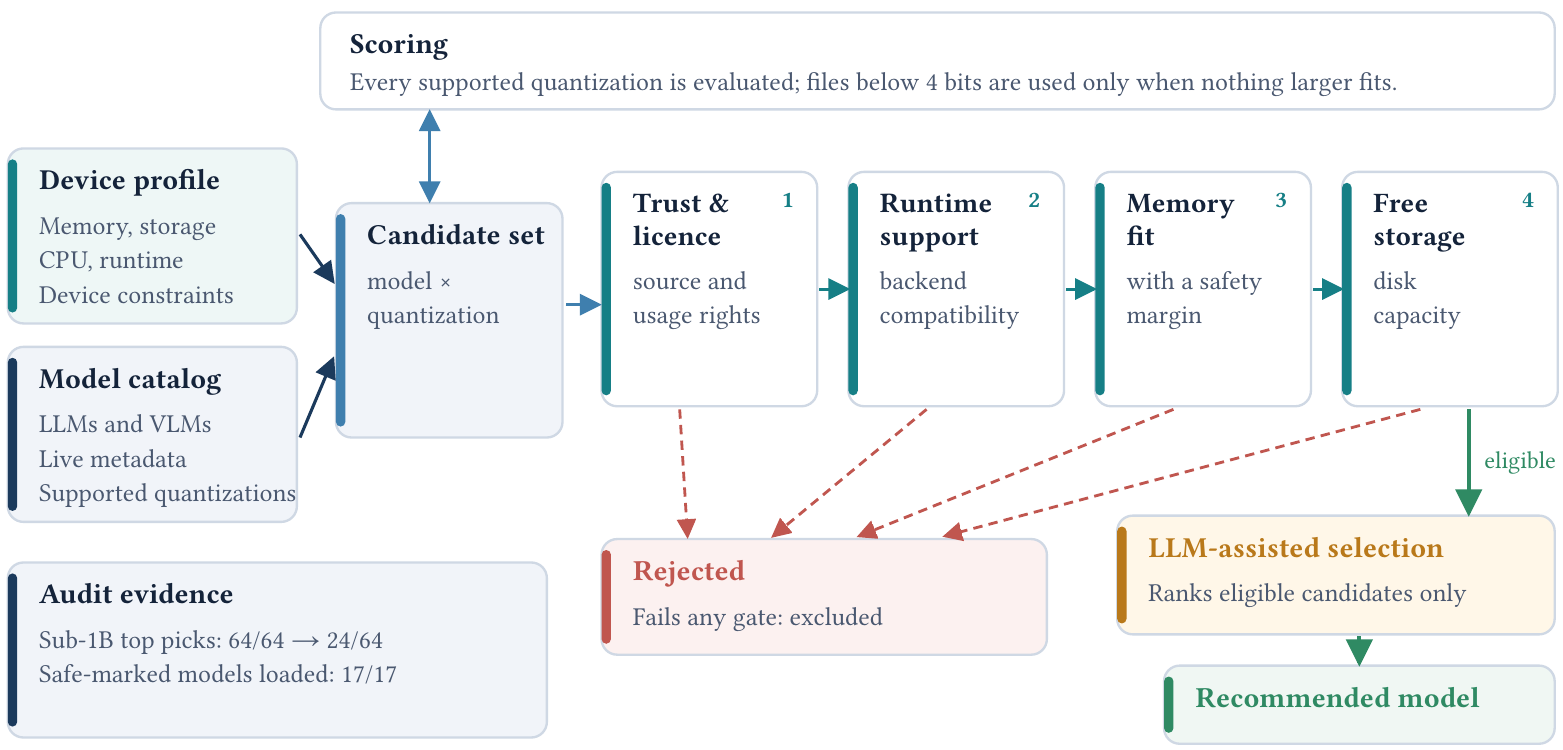}
\caption{Safety-aware model recommendation. Only candidates that pass every deterministic check reach agent-assisted selection. The audit evidence (lower left) is evaluated in Section~\ref{sec:fit}.}
\label{fig:safe}
\end{figure*}

The recommender combines language-model agents with deterministic checks, and the order in which they act is the core of the design (Figure~\ref{fig:safe}). The \emph{device profile} (memory, storage, processor, runtime and other constraints) and the \emph{model catalog} (language and vision-language models, live metadata from Hugging Face, supported quantizations) form a candidate set of model and quantization pairs. Each candidate must pass four gates in order: \textbf{trust and license} (source and usage rights), \textbf{runtime support} (backend compatibility), \textbf{memory fit} with a safety margin, and \textbf{free storage}. A candidate that fails any gate is excluded. Only the remaining candidates reach the agent, so the agent ranks but cannot override the checks.

Scoring evaluates every supported quantization of every model and uses files below 4 bits only when no larger option fits, following the frontier in Figure~\ref{fig:frontier}. The agent stage has four roles: a research agent that searches the web for current results, a reasoning agent that proposes a choice from the pre-scored options, a critic that checks the proposal against the scores and the device's memory budget, and an orchestrator that decides. Code maps the final choice to one of the pre-checked options, so an agent cannot introduce a model that the checks did not approve. The output is three recommendations with their optimizations, together with an explicit list of what the device should not run. The process is iterative: a user can override a choice or give feedback until the result is satisfactory.

\subsection{Optimization Pipeline}\label{sec:optimize}

Nexus does not stop at a recommendation. Selecting the model, its type and its level of quantization is a joint optimization over the device and the task, and once the user confirms a choice, Nexus performs the optimization. Quantization backends are available for llama.cpp files~\cite{llamacpp}, AWQ~\cite{awq}, GPTQ~\cite{gptq}, MLX~\cite{mlx} and BitNet~\cite{bitnet}. Fine-tuning with low-rank adapters~\cite{lora,qlora} adapts the model to the tasks the user cares about. Every download from Hugging Face is pinned to an inspected commit and checked against its published SHA-256 hash, and a file that fails the check is deleted.

\subsection{Native Device Clients}\label{sec:clients}

Nexus provides clients for Android, iOS and the desktop (Windows, Linux and macOS). A user installs the client and signs in; the client lists the optimized models available for that device, highlights the best fit, and downloads and runs the chosen model entirely on the device. Each client embeds a runtime suited to its platform: llama.cpp built into the app on Android, MLX or llama.cpp on iOS, and llama.cpp inside an Electron application on the desktop. The Android and desktop clients check the SHA-256 and size of every model that a remote job delivers. Before loading a model, the Android client compares the memory it needs (weights, cache and any vision add-on, plus headroom for the system) with the free memory and refuses if it does not fit. The desktop client reads the model's size from the file and the measured device memory, shortens the context if that helps, and otherwise refuses with a reason.

\subsection{Monitoring, Remote Evaluation and Feedback}\label{sec:monitor}

\begin{figure*}[t]
\centering
\includegraphics[width=\linewidth]{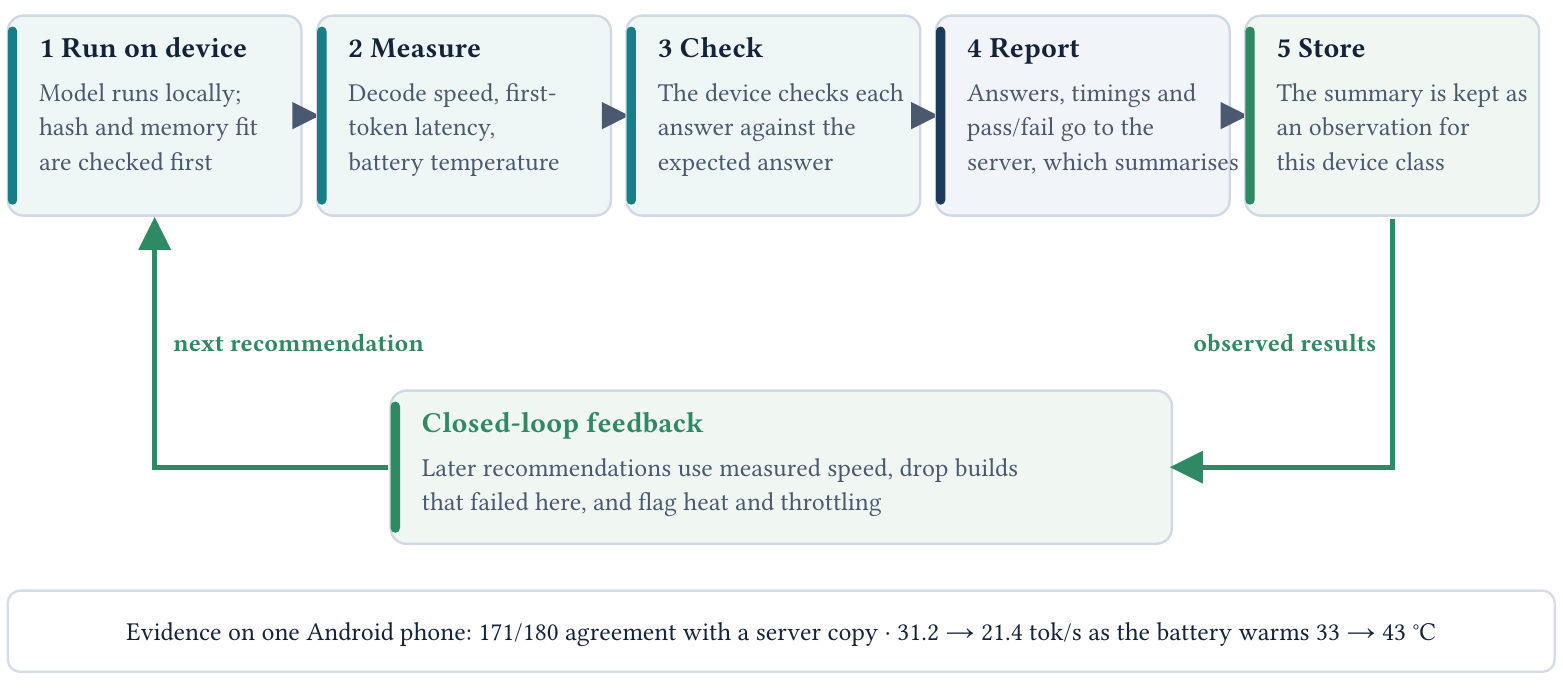}
\caption{Monitoring, remote evaluation and feedback. Device-side evidence closes the loop between deployment and the next model choice. The phone evidence at the bottom is evaluated in Section~\ref{sec:phone}.}
\label{fig:monitor}
\end{figure*}

A model that fits today may not remain the best choice. The device can lose capability over time (heat, battery wear, less free memory and storage), and newer models and better methods keep appearing. Nexus therefore keeps measuring after deployment, and the \emph{remote job} is the mechanism that turns a deployed device into a measurement instrument (Figure~\ref{fig:monitor}). The server, or a researcher, queues a job for a registered device; the job names a model, a list of prompts with optional expected answers and, optionally, engine settings such as thread count or context size. Phones ask the server for work at short intervals; the desktop client keeps a connection open. The device downloads and verifies the model, runs the prompts locally, and records decoding speed, time to first token and battery temperature from inside the native inference loop. It checks each answer against the expected answer (a substring or a pattern) and reports answers, timings and pass or fail to the server. The server summarizes the job (pass counts, speed, latency percentiles, peak temperature and a throttling ratio) and stores it as an observation. The Model Scout uses these observations in later recommendations: it drops a build that was refused or failed to load on that device, prefers the speed measured on that device or similar ones over its estimate, and lowers the rank of builds that answered few checks correctly, ran hot or slowed under sustained load. In our experiments a script also ran the same prompts on a server copy of the model to compare answers (Section~\ref{sec:phone}). Every phone result in Section~\ref{sec:eval} was produced through this mechanism.

\section{Evaluation}\label{sec:eval}

\subsection{Setup}\label{sec:setup}

\textbf{Testbeds.} We used three testbeds, ordered from most to least realistic. The \emph{real device} is one POCO F6 phone (Android 16, 8 GB of memory of which about 7 GB is available to apps, 256 GB of storage, 8 cores); every phone measurement ran on it as a remote job. The \emph{emulated devices} are server processes restricted to 1 to 8 pinned CPU cores and 2 to 16 GB of memory; they test fit logic and scaling with cores, but they have no heat, battery or phone memory system. The \emph{server} has an NVIDIA A100 GPU and a multi-core CPU, and serves as the reference for answer quality.

\textbf{Models.} We used models from 135M to 4B parameters, including SmolLM2~\cite{smollm2}, Qwen3 and Qwen3.5~\cite{qwen3}, LFM2.5 and SmolVLM~\cite{smolvlm}. Decoding is greedy (temperature 0) throughout: the model always picks its most likely next word, so no randomness comes from sampling. Answers can still differ slightly between devices, because different processors do the same arithmetic in a different order (Section~\ref{sec:phone}). Model files are named by their bits per weight (8-bit, 4-bit and so on); lower bits mean a smaller file and some loss of quality.

\textbf{Metrics.} \emph{Agreement} is the fraction of benchmark items that receive the same pass or fail on the phone and on the server. \emph{Perplexity} measures how well a model predicts held-out text (lower is better). \emph{Divergence} D is the Kullback--Leibler divergence~\cite{kl} of a compressed model's next-token distribution from the 16-bit model's on held-out text (lower is better); it separates files that a few hundred accuracy questions cannot (ARC-Easy varies by about $\pm$5 points). \emph{Decoding speed} is generated tokens per second, and \emph{time to first token} includes reading the prompt.

\subsection{Behavior on a Real Phone}\label{sec:phone}

\begin{findingbox}
\textbf{Finding 1.} The phone and the server give the same answers almost every time, but the phone writes more slowly as it warms and as the context grows, and it refuses models that leave too little free memory.
\end{findingbox}

\textbf{Answers agree.} For Qwen3.5-0.8B at 8 bits, 180 items from MMLU~\cite{mmlu}, ARC-Challenge~\cite{arc}, GSM8K~\cite{gsm8k} and HumanEval~\cite{humaneval} received the same pass or fail on the phone and on the server in 171 cases (Table~\ref{tab:phone}). The phone answered 70 items correctly and the server 73. Because the weights are identical and decoding is greedy, the remaining disagreements reflect the two processors performing the same arithmetic in a different order, which can tip a borderline answer either way. On a separate set of 24 checkable questions both got 22 right and missed the same two.

\begin{table*}[t]
\caption{Qwen3.5-0.8B at 8 bits on the phone and the server (same file, greedy decoding). \emph{Correct answers} counts items answered correctly; \emph{same pass/fail} counts items that both devices got right or both got wrong; speed is median decoding tokens per second (MMLU and ARC answers are a single letter, so their speed reflects one decoding step); battery is the phone's temperature during the task. Tasks are listed in the order they ran, so the phone warms down the table.}
\label{tab:phone}
\centering\small
\begin{adjustbox}{max width=\linewidth}
\begin{tabular}{lrrrrrrr}
\toprule
 & & \multicolumn{2}{c}{Correct answers} & Same & \multicolumn{2}{c}{Speed (tok/s)} & Battery \\
\cmidrule(lr){3-4}\cmidrule(lr){6-7}
Task & Items & Phone & Server & pass/fail & Phone & Server & (\textdegree{}C) \\
\midrule
MMLU & 60 & 24 & 22 & 58/60 & 31.2 & 55.0 & 33.1--34.1 \\
ARC-Challenge & 40 & 23 & 24 & 39/40 & 30.3 & 55.7 & 34.1--35.5 \\
GSM8K & 40 & 13 & 15 & 36/40 & 24.7 & 57.5 & 35.5--39.3 \\
HumanEval & 40 & 10 & 12 & 38/40 & 21.4 & 56.9 & 37.9--42.9 \\
\midrule
\textbf{Total} & \textbf{180} & \textbf{70} & \textbf{73} & \textbf{171/180} & & & \\
\bottomrule
\end{tabular}
\end{adjustbox}
\end{table*}

\textbf{Heat limits speed.} Over the same run, decoding speed fell from 31.2 to 21.4 tokens per second while the battery warmed from 33 to 43 \textdegree{}C, but that drop mixes heat with the tasks, whose answers grow from one token to a few hundred. A cleaner comparison uses identical settings and prompts: during a settings sweep, the default configuration ran at 26.0 tokens per second at the start and 15.4 when repeated midway, with the battery at about 39 and 41 \textdegree{}C. A single speed figure therefore misrepresents a phone. Nexus records the temperature with every measurement so that it can report a cool and a warm speed for each device.

\textbf{Longer context slows writing.} Chat users notice that a model writes more slowly as a conversation grows. We measured this on the phone with Qwen3.5-0.8B at 4 bits: prompts of 35 to about 1,200 tokens, each followed by 64 generated tokens, in a shuffled order and with a short prompt after every third one. Comparing each long prompt with the short prompts run next to it removes the effect of warming. Writing speed fell to 88\% of the short-prompt speed at 100 to 300 tokens of context and to 82\% at about 1,200 tokens, roughly 7--8\% per 1,000 tokens. Reading the prompt ran at 60 to 74 tokens per second, so a 1,200-token prompt waited about 18 seconds for its first word.

\textbf{The memory guard holds in practice.} The 2.6B LFM2.5 model at 4 bits needs about 2.5 GB in total by the client's estimate (weights, cache and a small allowance), and the client also keeps 450 MB free for the system. With other applications open, the phone had only 2.7 to 2.9 GB free, which is below the client's safety margin, so the client refused to load the model. Its 1.2B sibling (661 MB) loaded and decoded at about 21 tokens per second with the battery between 31 and 38 \textdegree{}C. The practical ceiling of an 8 GB phone in daily use is therefore below 2.5 GB for the model and its cache.

\textbf{Vision runs on the phone.} Three vision models answered the same 40 drawn pictures with known answers (the color of a shape, the number of circles, a printed word, which side holds the larger shape, and the name of a shape) on the phone and on the server (Table~\ref{tab:vision}). SmolVLM-500M answered 39 correctly on both and agreed on all 40; Qwen3.5-2B answered 38 on the phone and 37 on the server; Qwen3.5-0.8B answered 32 and 30 and agreed on 38, with most of its misses in counting circles. How long the phone takes to read a picture varies widely between models and depends not just on parameter size but also on the architecture of the vision component: Qwen3.5-0.8B needed about 5 seconds, SmolVLM-500M and Qwen3.5-2B 18 to 21 seconds, and the server 1.5 to 2. Repeating the SmolVLM-500M test five days later, on a newer version of the app, gave the same answer to all 40 pictures.

\begin{table}[t]
\caption{Vision models on the phone and the server: 40 drawn pictures with known answers. Size is the model plus its vision component in MB; \emph{read} is the median time on the phone from picture to first word, and tok/s its writing speed.}
\label{tab:vision}
\centering\small
\begin{adjustbox}{max width=\linewidth}
\begin{tabular}{lrrrrrr}
\toprule
Model & MB & Phone & Server & Agree & Read (s) & tok/s \\
\midrule
SmolVLM-500M & 636 & 39 & 39 & 40/40 & 20.9 & 62.5 \\
Qwen3.5-0.8B & 698 & 32 & 30 & 38/40 & 5.3 & 29.9 \\
Qwen3.5-2B & 1841 & 38 & 37 & 37/40 & 18.3 & 14.9 \\
\bottomrule
\end{tabular}
\end{adjustbox}
\end{table}

\subsection{Safety on Emulated Devices}\label{sec:fit}

\begin{findingbox}
\textbf{Finding 2.} In our tests, every model the recommender marked safe loaded successfully, and the client's own guard never let an unsafe model run. The current recommender also chooses larger models when they fit, instead of always the smallest.
\end{findingbox}

\textbf{The verdicts are safe.} Five Qwen3 builds ran on eight emulated devices, giving 40 device and model pairs, under a strict memory limit that also counts the model file held in memory (a lenient limit that does not count it gives the other bound). The recommender gives each pair one of four labels: ``recommended'', ``fits'', ``tight'' (close other apps first) or ``won't fit''. All 17 pairs labeled ``recommended'' or ``fits'' loaded successfully; of the 8 labeled ``tight'', 3 loaded and 5 failed, as the warning intends; and 14 of the 15 labeled ``won't fit'' did fail. The unmodified desktop client, run against the live server with each emulated device reporting its own memory and cores, agreed with the real outcome in all 15 of its cases: it ran the 9 models that run and refused the 6 that fail.

\textbf{The recommender balances model size and precision.} Our first scorer always took the highest precision that fitted. That sounds safe, but it meant small models at high precision won almost every time: across 64 scenarios it chose a model under one billion parameters in all 64. The current scorer compares every precision of every model and keeps the best quality that fits, going below 4 bits only when nothing larger fits. It now finds a balance between size and precision: it chooses a model under one billion parameters in 24 of 64 scenarios, when the device is genuinely small, and larger compressed models elsewhere (Table~\ref{tab:audit}). The catalog grew from 36 to 58 models between the two audits, so we also restricted both scorers to the same fixed set of benchmark models: the first scorer picked the smallest of them in 56 of 64 scenarios, the current one in 8.

\begin{table}[t]
\caption{The same audit run on the first scorer and on the current one.}
\label{tab:audit}
\centering\small
\begin{adjustbox}{max width=\linewidth}
\begin{tabular}{lrr}
\toprule
Measure & First & Current \\
\midrule
Sub-1B top pick (64 scenarios) & 64/64 & 24/64 \\
Smallest of a fixed model set picked & 56/64 & 8/64 \\
Catalog size (LLM + VLM) & 36 & 58 \\
\bottomrule
\end{tabular}
\end{adjustbox}
\end{table}

\subsection{Compression on the Server}\label{sec:compress}

\begin{findingbox}
\textbf{Finding 3.} Compression costs some quality, but a few chunks of calibration text recover most of it. For a fixed amount of memory, the best choice is the largest model at 3--4 bits.
\end{findingbox}

\textbf{Quality follows file size, with exceptions.} For Qwen3.5-0.8B in twelve llama.cpp formats, 8-bit and 6-bit files stayed within 1.3\% of the full-precision perplexity, and storing the model's working memory (its key--value cache) at 8 bits cost only 0.17\% while halving that memory. Two-bit files were unusable without an importance matrix: perplexity rose by 472\%. Writing speed on the server CPU varied by less than a third across formats, from 59.8 tokens/s at 8 bits to 78.6 at 2 bits. The scorer's memory estimates were too low for low-bit files, and the error grew as bits fell (10 to 18\% at 4 bits).

\textbf{4-bit formats cost a few points of accuracy.} For SmolLM2-1.7B-Instruct, the 16-bit model scored 71.0\% on ARC-Easy~\cite{arc}; GPTQ lost 5.7 points, AWQ 7.0 and NF4 8.3 (Table~\ref{tab:formats}). With 300 questions the 95\% interval is about $\pm$5 points, so accuracy alone cannot rank the three 4-bit formats; perplexity on WikiText-2~\cite{wikitext} ranks them AWQ, NF4, GPTQ. The server has no C compiler, which AWQ's fast kernel needs, so AWQ was evaluated with unpacked weights (its quality, not its speed), and GPTQ ran on a slow fallback kernel at 14.9 tokens/s against 45.1 for 16-bit.

\begin{table}[t]
\caption{SmolLM2-1.7B-Instruct in four precisions on the A100 (ARC-Easy, 300 questions; WikiText-2 perplexity).}
\label{tab:formats}
\centering\small
\begin{adjustbox}{max width=\linewidth}
\begin{tabular}{lrrr}
\toprule
Format & MB & ARC-E (\%) & PPL \\
\midrule
BF16 & 3264 & 71.0 & 11.36 \\
NF4 & 1056 & 62.7 & 12.97 \\
AWQ & 990 & 64.0 & 12.16 \\
GPTQ & 993 & 65.3 & 13.61 \\
\bottomrule
\end{tabular}
\end{adjustbox}
\end{table}

\textbf{An importance matrix is the cheapest gain.} An importance matrix, computed from a few chunks of ordinary text, tells the quantizer which weights matter most. For SmolLM2-1.7B it lowered D from 0.118 to 0.036 at 4 bits (1.06 GB) and from 1.01 to 0.47 at 2 bits, at identical file size (Table~\ref{tab:imatrix}). Five chunks of 512 tokens worked as well as 400; text from the wrong domain cost at most 7\%. With the matrix, llama.cpp 4-bit files were 3 to 6 times closer to the original model than AWQ, NF4 and GPTQ files of the same model (D of 0.036 against 0.105, 0.164 and 0.227).

\begin{table*}[t]
\caption{Calibration makes compressed files much closer to the original. Each value is D for SmolLM2-1.7B: how far the compressed model's next-word predictions drift from the 16-bit model's, on held-out prose and code (lower is better). \emph{Plain} files are compressed directly; \emph{calibrated} files use an importance matrix computed from a few chunks of ordinary text. At every precision, calibration cuts the drift by roughly half to three quarters (49 to 74\%) at the same file size, and calibrated 4-bit files are closer to the original than the three GPU 4-bit formats in the last row.}
\label{tab:imatrix}
\centering\small
\begin{adjustbox}{max width=\linewidth}
\begin{tabular}{lrrrrr}
\toprule
Precision & MB & Prose, plain & Prose, calibrated & Code, plain & Code, calibrated \\
\midrule
4-bit & 1056 & 0.118 & 0.036 & 0.100 & 0.026 \\
4-bit, compact & 940 & 0.125 & 0.057 & 0.099 & 0.041 \\
3-bit & 860 & 0.250 & 0.128 & 0.203 & 0.091 \\
2-bit & 675 & 1.010 & 0.465 & 0.844 & 0.360 \\
AWQ / NF4 / GPTQ &  & \multicolumn{4}{l}{0.105 / 0.164 / 0.227 (prose); 0.072 / 0.131 / 0.166 (code)} \\
\bottomrule
\end{tabular}
\end{adjustbox}
\end{table*}

\begin{figure}[t]
\centering
\includegraphics[width=\linewidth]{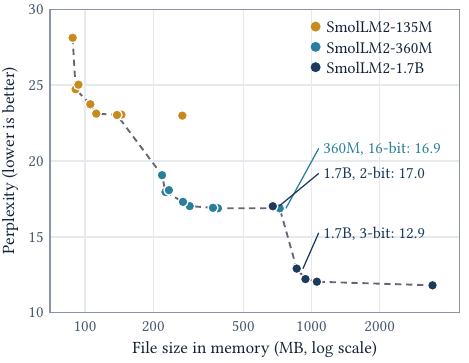}
\caption{Quality against memory for one model family. Each point is a real llama.cpp file of SmolLM2 at one size and precision (16 down to 2 bits per weight); perplexity is on held-out prose. The dashed line is the Pareto frontier. Down to about 3 bits, a larger compressed model beats a smaller one at full precision (Section~\ref{sec:compress}).}
\label{fig:frontier}
\end{figure}

\textbf{Spend memory on parameters, down to about 3 bits.} Figure~\ref{fig:frontier} shows the result that guides the recommender. At a budget of about 0.9 GB, SmolLM2-1.7B at 3 bits (860 MB) reaches a perplexity of 12.9, while the 360M model at 16 bits (724 MB) reaches only 16.9; at 270 MB the 360M model at 4 bits (17.3) beats the 135M model at 16 bits (23.0). Below 3 bits the advantage disappears: the 1.7B model at 2 bits (17.0) merely ties the 360M model at full precision. This agrees with scaling studies of precision~\cite{kbit} and with on-device measurements that place the threshold near 3.5 bits per weight~\cite{song2025}, and gives Nexus a default: prefer the largest model that fits at 3--4 bits with an importance matrix, rather than a smaller model at 6 or 8 bits.

\textbf{A vision model must receive the picture.} We compared three ways of answering a question about a picture with SmolVLM-256M, on 30 generated pictures per task (Table~\ref{tab:native}): sending the picture itself, sending only text extracted from it (perfect extraction), and sending no picture at all. Sending the picture scored 100\% on naming a color, 53\% on counting circles and 100\% on reading a word. Text only scored 0\%, 0\% and 100\%, and no picture scored 0\%, 33\% and 0\%. Text extraction therefore suffices for reading words but fails on anything visual, and the picture's advantage is smallest in counting: 53\% against 33\% is 16 against 10 correct answers out of 30. Nexus therefore passes the picture to a vision model and, when the loaded model has no vision projector, offers to switch to a vision model that fits the device's free memory, fetching and hash-checking both files.

\begin{table}[t]
\caption{SmolVLM-256M: the picture itself against text extracted from it (30 pictures per task, accuracy in \%).}
\label{tab:native}
\centering\small
\begin{adjustbox}{max width=\linewidth}
\begin{tabular}{lrrr}
\toprule
Task & Picture & Text only & None \\
\midrule
Color & 100 & 0 & 0 \\
Count & 53 & 0 & 33 \\
Word & 100 & 100 & 0 \\
\bottomrule
\end{tabular}
\end{adjustbox}
\end{table}

\subsection{Using the Capacity a Model Leaves Free}\label{sec:spare}

A small model leaves most of a phone's memory idle, and generating text is largely waiting for the weights to be read from memory. If several requests share each read, total output can rise. We measured this on the server restricted to four pinned cores, which stand in for a phone, with continuous batching as in Orca~\cite{orca}, as implemented in llama.cpp~\cite{llamacpp}, 96 tokens per request and three interleaved repetitions (Table~\ref{tab:batch}). At four requests, total throughput was 1.71 to 1.88 times that of one request, and eight requests gave no further gain. Each user then sees roughly half the single-request speed. These are server cores, not a phone; the phone also slows when hot, so the gain must be re-measured there at steady temperature.

In our measurements on four pinned cores, writing speed tracked memory bandwidth: across 11 llama.cpp formats, nine had a product of file size and writing speed between 33 and 41 GB/s, while the two most compact formats, below 3 bits, reached only 13 to 15 GB/s probably because unpacking their weights costs more than reading them. Extra threads therefore help reading the prompt (29, 57, 109, 212 and 264 tokens/s for 1, 2, 4, 8 and 16 threads) but not writing. With prompt caching on, the first word of the tenth chat turn arrived after 0.85 s instead of 7.7 s (1.7B, four cores), and restoring a saved conversation state took 99 ms against 38.8 s to read a 3,483-token history again. Speculative decoding~\cite{specdec} with a 135M draft model sped up writing by 41\% for JSON and 30\% for code but slowed chat by 21\%.

\begin{table}[t]
\caption{Total throughput with several requests served together, relative to one request (four pinned cores, 96 tokens per request, three repetitions).}
\label{tab:batch}
\centering\small
\begin{adjustbox}{max width=\linewidth}
\begin{tabular}{lrrrr}
\toprule
Model & 1 req (tok/s) & 2 & 4 & 8 \\
\midrule
SmolLM2-135M & 259.6 & 1.33$\times$ & 1.71$\times$ & 1.72$\times$ \\
LFM2.5-2.6B & 20.2 & 1.14$\times$ & 1.88$\times$ & 1.77$\times$ \\
Qwen3.5-4B & 11.3 & 1.24$\times$ & 1.75$\times$ & 1.57$\times$ \\
\bottomrule
\end{tabular}
\end{adjustbox}
\end{table}

\subsection{What Did Not Help}\label{sec:nohelp}

\textbf{Auto-tuning engine settings on the phone.} A search over thread count, batch size and flash attention~\cite{flashattn} found no setting that beat the defaults beyond noise. A first single-pass sweep was misleading, because the default configuration itself slowed from 26.0 to 15.4 tokens per second as the phone warmed during the sweep. We now alternate each setting with the default, shuffle the order, and compare each setting only with its neighboring defaults.

\textbf{More threads for reading the prompt.} Giving the phone more threads to read the prompt looked 11--18\% faster in a single pass. But the phone warms during a long test, which slows later settings and flatters earlier ones. When we alternated each setting with the default so that warming affected both equally, the difference disappeared: both reading and writing stayed within 2\% of the default.

\textbf{Per-layer bit allocation.} Assigning bit widths per layer by measured sensitivity improved quality in a simulator (math loss 1.277 to 1.198 at the memory of uniform 4-bit), but the advantage mostly vanished when we built real model files. We keep the sensitivity map as a ranking signal and do not ship the allocation.

\section{Conclusion}\label{sec:conclusion}

Nexus brings AI to everyday devices by treating on-device deployment as a closed loop with safety as its first constraint. Our current work concentrates on phones, because they are the devices most people use every day, and on language and vision-language models. Within the limits below, the evidence supports the design: deterministic checks kept unsafe models off devices, the largest model at 3--4 bits made the best use of limited memory, and devices that tested their own models produced results that agreed with the server and showed how real use slows a phone.

\section{Limitations and Future Work}\label{sec:limits}

All phone results come from a single phone, so they describe a method rather than a population of devices; the first step is to repeat them on a second phone, and then on iOS devices and Windows laptops. Emulated devices reproduce memory limits and core counts but not heat, battery behavior or a phone's memory system, the vision tests use drawn pictures rather than photographs, and context was tested only up to about 1,200 tokens. Several studies use small samples, so differences of a few points should not be over-interpreted. The device damage that motivated our safety design (Section~\ref{sec:intro}) is a single observation that we have not reproduced. Beyond phones, we want to extend Nexus to IoT and edge devices, which usually have no screen and need a client of their own, to let a deployed model keep learning its user's tasks without forgetting earlier ones, and to test our agents against indirect prompt injection~\cite{injection}.

\bibliography{refs}

\bibliographystyle{icml2025}
\end{document}